\documentclass[conference]{IEEEtran}
\input epsf
\usepackage{graphicx}
\usepackage{pgfplots}
\usepackage{multirow}
\begin{document}

\title{\LARGE OCR for TIFF Compressed Document Images Directly in Compressed Domain Using Text segmentation and Hidden Markov Model}

\author{\IEEEauthorblockN{Dikshit Sharma \textsuperscript{1}, Mohammed Javed \textsuperscript{2}}
\IEEEauthorblockA{\textit{Department of IT,}
\textit{Indian Institute of Information Technology Allahabad,} India 211015}
\textit{\textsuperscript{1}sharma.dikshit001@gmail.com} 
\textit{\textsuperscript{2}javed@iiita.ac.in} }


\maketitle

\begin{abstract}
In today's technological era, document images play an important and integral part in our day to day life, and specifically with the surge of Covid-19, digitally scanned documents have become key source of communication, thus avoiding any sort of infection through physical contact. Storage and transmission of scanned document images is a very memory intensive task, hence compression techniques are being used to reduce the image size before archival and transmission. To extract information or to operate on the compressed images, we have two ways of doing it. The first way is to decompress the image and operate on it and subsequently compress it again for the efficiency of storage and transmission. The other way is to use the characteristics of the underlying compression algorithm to directly process the images in their compressed form without involving decompression and re-compression. In this paper, we propose a novel idea of developing an OCR for CCITT  \textit {(The International Telegraph and Telephone Consultative Committee)} compressed machine printed TIFF document images directly in the compressed domain. After segmenting text regions into lines and words, HMM is applied for recognition using three coding modes of CCITT- horizontal, vertical and the pass mode. Experimental results show that OCR on pass modes give a promising results. 
\\
\newline
\end{abstract}
\IEEEoverridecommandlockouts

\IEEEpeerreviewmaketitle

\section{Introduction}
\begin{figure}
      \includegraphics[width=3.7in]{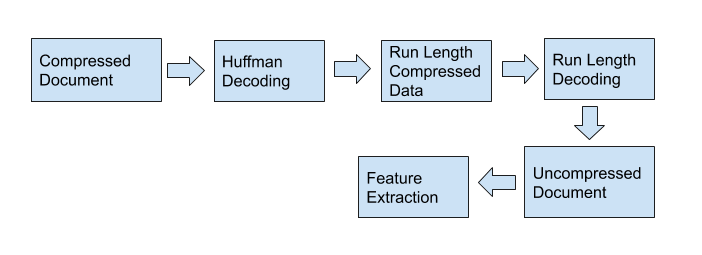}
     \caption{Classical method of feature extraction from compressed images involving decompression }
     \label{Figure1}
 \end{figure}

Document Image Analysis (DIA) \cite{c1,c3} is a technique like any other digital image analysis that takes scanned document images as input, and performs major operations like feature extraction, segmentation and recognition, in order to accomplish the dream of moving towards paperless office. One major problem with scanned document images is that they occupy large storage space for archival and high bandwidth for transmision \cite{c1, c3}. Therefore, in the literature, various compression techiques have been evolved to make the task of storage and archival very economical. Image acquisition devices now come with default compression algorithms, as a result, in the real world most of the images are made available in the compressed form. Now in order to operate with these compressed images there are two ways, the first way is to decompress the image and operate over it, and subsequently compress it again for the efficiency of storage and transmission which is termed as \textit{conventional image analysis} \cite{c13}. The other way is to use the characteristics of the underlying compression algorithm to directly process the images in the compressed form without using decompression and re-compression stages is called as \textit{compressed domain analysis}. Many interesting research works have been reported in the field using both handcrafted features and deep learning features, all executed directly in the compressed representations  \cite{c3,c14,c15,c16,c17,c18,c19,c20}. The different stages involved in a typical conventional image processing (case of feature extraction) involving decompression is illustrated through Figure-\ref{Figure1}. The aim of this research paper is to accomplish feature extraction, segmentation and Optical Character Recognition (OCR) directly in compressed document images without involving decompression and re-compression operations. 

Although a lot of work has been reported in developing OCRs for uncompressed images/documents \cite{c13}, not much work has been carried out in realizing OCRs for
compressed document images. The initial idea to work with the compressed
documents images directly was thought off by some researchers in the
early 1980’s \cite{c2}. There are several efforts being made to directly handle the
images intelligently in the compressed form \cite{c3}. Some of those operations
are feature extraction, segmentation, image rotation, skew detection, connected component analysis etc \cite{c3}.
In all these operations either the run length information is used from the
uncompressed form or either some sort of partial decoding is done to carry out the operations. In this work we present a system that will recognize the text directly from the CCITT compressed TIFF images \cite{c1}. 

The OCR model proposed in this paper is similar to that of \cite{c2} that extracts the feature points marked by the pass modes directly from the compressed file using Hidden Markov Models (HMMs) as recognizer. HMMs become an obvious choice when we are dealing with data comprising of a lot of noise. The pass mode features extracted from the compressed file are similar to noise. HMMs perform better in the presence of noise and hence they are extensively used in speech and handwriting recognition. However, this research paper fulfills two research gaps mentioned in \cite{c2}. The first contribution here is to accomplish the novel idea text segmentation into lines and words directly in compressed text using three modes- horizontal, vertical and pass modes of CCITT Group 4 compression. The second contribution here is using three modes for realising OCR, and further experimentally proving that pass mode is better for OCR and other document image related operations to be performed in the compressed domain. 
Rest of the paper is organized as follows- Section II gives brief idea of the proposed model with feature extraction, text line segmentation and HMM, section III reports experimental results and related analysis, and finally section IV summarizes the research paper.
 
\section{Proposed Model}
\begin{figure*}[h]
      \includegraphics[width=\textwidth]{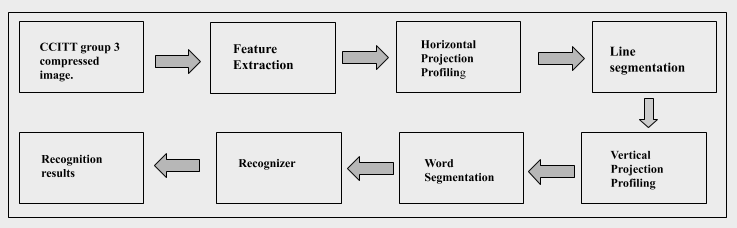}
     \caption{Different stages in the proposed OCR for TIFF Compressed Document Images}
     \label{Figure2}
    \end{figure*}

The OCR proposed in this paper follows three major stages - feature extraction, text segmentation and OCR using HMM. The different stages are illustrated through the Figure-\ref{Figure2}.
      
\subsection{Feature extraction}

The feature extraction module is used to extract feature points by handling the images directly in the compressed domain. Now as we are dealing with the compressed images so it is pretty evident that the feature points shall be sparse, so, the pass codes are extracted from: left to right / top to bottom, right to left / bottom to top.
Thus in two different ways the feature points are extracted and overlaid on top of one another and given as an input to the segmentation module which will perform projection profiling on it and then finally it is given as an input to the recognition module.
Our recognition module is based on Hidden Markov Model (HMM) \textit{discussed in detail in section 5}. It is a probabilistic model based on Markov Chains which computes the hidden state based on the state that is present i.e. the state that can be observed by making inferences(probabilistic inferences). 
    
The feature extraction module is used to extract feature points by handling the images directly in the compressed domain.Now as we are dealing with the compressed images so it is pretty evident that the feature points shall be sparse, so, the pass codes are extracted from: Left to right / Top to bottom and Right to left / Bottom to top. Thus in two different ways the feature points are extracted and overlaid on top of one another. This feature sequence is then given to the recognizer as input. Now as per the \textit {CCITT} group 3 algorithms the image can be compressed in two different ways, i.e., either using only a one-dimensional structure of the image or by using the two-dimensional information. In one-dimensional compression each line is treated individually on the other hand in two-dimensional compression the previous line is also relevant.In the following two subsections the both the coding schemes are briefly reviewed. Then in subsection II(3) \& and II(4) feature extraction and the line and word segmentation on the feature set is briefly described. 
 \subsubsection{Group 3 One Dimensional Coding}
    As the name suggests the Group 3 One Dimensional algorithm processes the image line by line from top to bottom. While processing each line the number of consecutive pixels of the same colour (black or white) are counted and a sequence of numbers or run-lengths are generated which represent a line (shown in Fig. 3.1). Each transition in the pixels in the image causes an entry in the coded image. Now to make the representation unique, it is assumed that each line begins with a white pixel. If a line actually starts with a black pixel than the number of leading white pixels is zero. This coding scheme is called run-length coding. \\
    Now this simple coding scheme has both its pros and cons. If there are long homogeneous runs of same colored pixels than only few values are needed to code the image and heavy compression can be achieved but on the other hand if there are frequent transitions between black and white many values shall be needed. So, in the worst case no compression may be achieved at all. Because in real life not all numbers of run-lengths occur with same probability , an additional Huffman coding is used for further compression.
    
    
    \begin{figure}[h!]
        \begin{center}
             \includegraphics[width=3.5in]{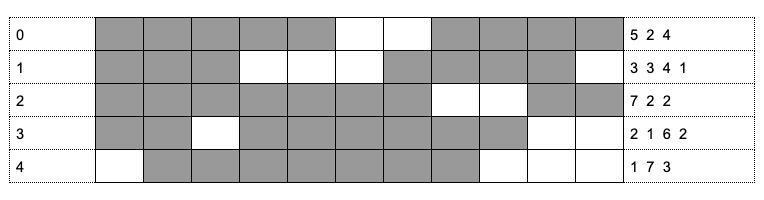}
            \caption{Group 3 One Dimensional Coding}
        \end{center}
     
    \end{figure}
    
    \subsubsection{Group 3 Two Dimensional Coding}
    In Group 3 One Dimensional coding where each line is treated independently without considering the previous and the following line. But in almost all cases, images lines depend on their context, i.e., the preceding and the following lines. This fact can be used for more elaborated coding techniques.
    In the CCITT group 3 two-dimensional coding scheme both the preceding and the following lines are considered for both compression as well as decompression. In this coding scheme five points are of special interest :$a_0$  the last known pixel in the actual line, $a_1$ the first transition pixel to the right of $a_0$,  $a_2$ the second transition pixel to the right of $a_0$,  $b_1$ on the previous line the first transition pixel to the right of $a_0$ and finally $b_2$ on the previous line the first transition pixel to the right of $a_0$.
        
Now depending on the position of the above points three coding modes are defined :
    \begin{enumerate}
        \item The horizontal mode, when $|a_1 - b_1| > 3$
        \item The vertical mode, when $|a_1 - b_1| <= 3$
        \item And the pass mode, when $|a_1 - b_2| > 0$
    \end{enumerate}
    
    \begin{figure}[h!]
        \begin{center}
            \includegraphics[width=3.5in]{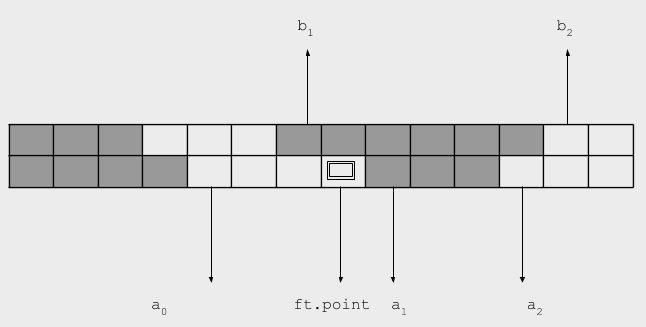}
            \caption{Group 3 Two Dimensional Coding}
        \end{center}
    \end{figure}
    
    Now the horizontal mode represents one-dimensional run length coding, the vertical and the pass modes concerns the structure of the previous line. In pass modes $b_1$ \& $b_2$ are are on the left side of $a_1$. To avoid the use of large number for coding the run length between $a_0$ \& $a_1$, a special mark is set right below $b_2$. This allows to reconstruct the colour of the pixels in the actual line to the position below $b_2$, which has the same colour as $a_0$. The position of this special mark is called pass code.\\
    Extraction of these points is quite straightforward. Figures 6, 7, 8 shows the feature points marked by horizontal, vertical \& pass modes respectively for the input image shown in Figure 5.
    
    \begin{figure}[h!]
        \begin{center}
             \includegraphics[width=3.5in]{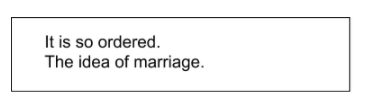}
            \caption{Sample input image for illustration purpose}
        \end{center}
     
    \end{figure}
    \begin{figure}[h!]
        \begin{center}
             \includegraphics[width=3.5in]{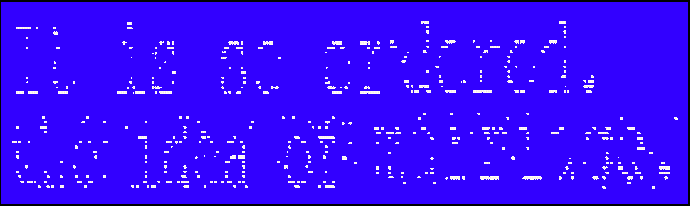}
            \caption{Feature points marked by horizontal mode for Fig 5}
        \end{center}
     
    \end{figure}
    \begin{figure}[h!]
        \begin{center}
             \includegraphics[width=3.5in]{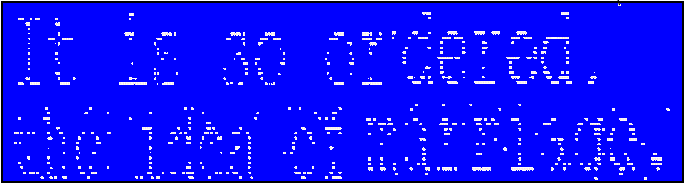}
            \caption{Feature points marked by vertical mode for Fig 5}
        \end{center}
    \end{figure}
    \begin{figure}[h!]
        \begin{center}
             \includegraphics[width=3.5in]{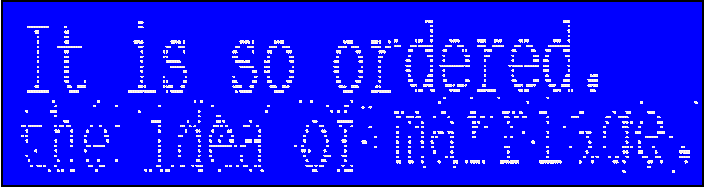}
            \caption{Feature points marked by pass mode for Fig 5}
        \end{center}
    \end{figure}

\subsection{Text line \& word segmentation using pass modes}
The output of the feature extraction module is then given as an input to the segmentation module to  perform profiling. A projection profile is a histogram of the number of set pixels along the parallel lines and horizontal lines. On the basis of that we shall perform segmentation both word and line segmentation. For an image consisting of 'm' rows and 'n'  columns, mathematically we can define the Horizontal Projection Profile (HPP) and  Vertical Projection Profile (VPP) as: \\
    $$ VPP(y) =\sum _{1<=x<=m}f(x,y);
    HPP(x) =\sum_{1<=y<=n}f(x,y)$$
    
    Now the run histograms generated by projection profiling can be used to analyze our feature set, like the amount of blank spaces available within the input i.e. between the lines and words in the document. The same has been demonstrated in the figures 11 and 12 where the blue lines represent the run of pixels and the red line representing the break between the respective lines/words for the input set and feature set shown in Fig 9 and Fig 10 respectively.
     
    \begin{figure}[h!]
        \begin{center}
             \includegraphics[width=3.5in]{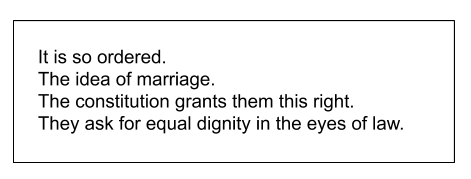}
            \caption{Input image}
        \end{center}
    \end{figure}
    
    \begin{figure*}[h!]
        \begin{center}
             \includegraphics[width=\textwidth, height=1.9in]{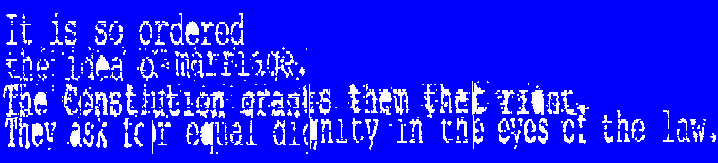}
            \caption{Feature set}
        \end{center}
    \end{figure*}
    
     Now after this much pre-processing the feature set shall be given as an input to our recognition module.
    
    \begin{figure}[h!]
        \begin{center}
             \includegraphics[width=3.5in]{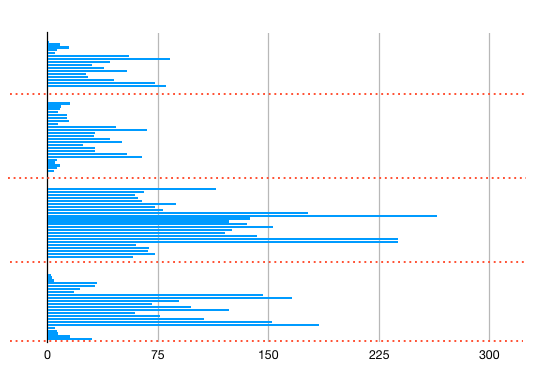}
            \caption{Sample text line segmentation for Fig 9}
        \end{center}
    \end{figure}
    
    \begin{figure}[h!]
        \begin{center}
             \includegraphics[width=3.5in]{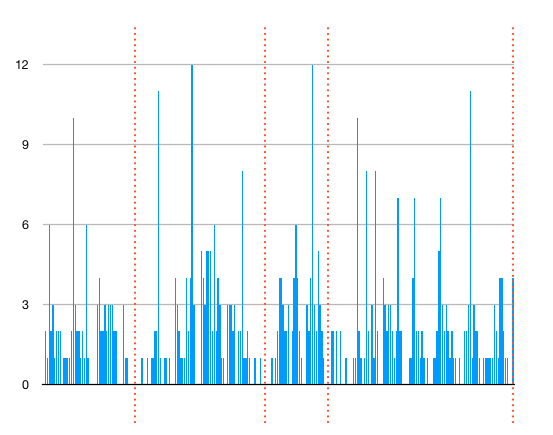}
            \caption{Word segmentation for a single text line from Fig 9}
        \end{center}
    \end{figure}

\subsection{Recognition Module}
    In this section we shall discuss what Hidden Markov Models are and how we used them to do OCR. 
    \subsubsection{Markov Chains}
         Hidden Markov Models (HMMs) are elicited from a concept of mathematics called Markov Chains. Markov Chain is a probabilistic model that helps us compute the probabilities of a sequence of random variables which takes up values from some pre-defined sets, these pre-defined sets can be anything like symbols, colors, text etc. The sate in which the system is presently at it makes an prediction/assumption of the future state. So, the states preceding the present state have no influence on the prediction e.g on the basis of the even that's happening right now you have to predict the future event without having access to any other information like the events that have preceded the current event.
         \begin{figure}[h!]
        \begin{center}
            \includegraphics[width=3.5in]{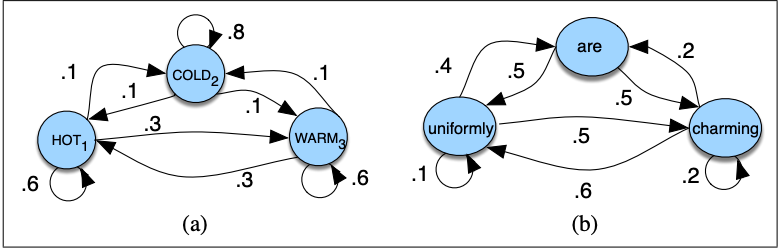}
            \caption{Markov Chain example}
         \end{center}
    \end{figure}
    Figure $13$ contains two different examples of the Markov Chains 13(a) for weather and 13(b) for words clearly showing different states and transitions between them. Now a starting probability distribution is required $\pi = [0.2, 0.4, 0.4]$ for $(a)$ and this would mean that 0.2 is probability of starting with HOT, and both COLD and WARM have a probability of 0.4 to start with. And rest of them are the transition probabilities i.e., the probabilities of going from one state to another. \\
    It consists of the following components : 
    \begin{enumerate}
        \item $Q = q_1, q_2, q_3, .. , q_n$  \\A set of \textbf{N} states.
        \item $T = t_{11}, t_{12}, t_{13}, .. , t_{nm}$\\
        A transition probability matrix T, each $t_{ij}$ represents the probability of transitioning from state $i$ to state $j$ such that 
        $\sum_{i = 1} ^ {n} t_{ij} = 1 \qquad \forall i$ .
        \item $\prod = \pi_1, \pi_2, .. , \pi_n$ \\
        Set of initial probability distribution representing the probability of the Markov Chain to start with a specific state. Also $\sum_{i = 1} ^{n} = 1$.
    \end{enumerate}
    \subsubsection{Hidden Markov Models}
      Now when we need to predict something on the basis of what we observe a Markov Chain seems to be an possible choice e.g. on the basis of the weather outside we wish to predict the mood of someone. We call this event hidden because this cannot be observed directly. \\
    Now \textit{Hidden Markov Model} is a probabilistic model based on \textit{Markov Chains} that allows us to figure out these hidden states. A \textit{Hidden Markov Model} comprises of :
    \begin{enumerate}
        \item $Q = q_1, q_2, q_3, .. , q_n$  \\A set of \textit{N} states.
        \item $T = t_{11}, t_{12}, t_{13}, .. , t_{nm}$\\
            A \textit{transition probability matrix T}, each $t_{ij}$ represents the probability of transitioning from state $i$ to state $j$ such that 
            $\sum_{i = 1} ^ {n} t_{ij} = 1 \qquad \forall i$ .
        \item $\prod = \pi_1, \pi_2, .. , \pi_n$ \\
            Set of \textit{initial probability distribution} representing the probability of the Markov Chain to start with a specific state. Also $\sum_{i = 1} ^{n} = 1$.
        \item  $O = o_1, o_2, o_3, .. , o_m$  \\A sequence of \textit{M} observations.
        \item $B = b_i(o_t)$ \\
        Set of emission probabilities.
    \end{enumerate}
    Figure $14$ shows a sample HMM for the ice cream task. The two hidden states
    (H and C) correspond to hot and cold weather, and the observations (drawn from the alphabet $O = {1,2,3})$ correspond to the number of ice creams eaten by the person on a given day.
     \begin{figure}[h!]
            \begin{center}
                \includegraphics[width=3.5in]{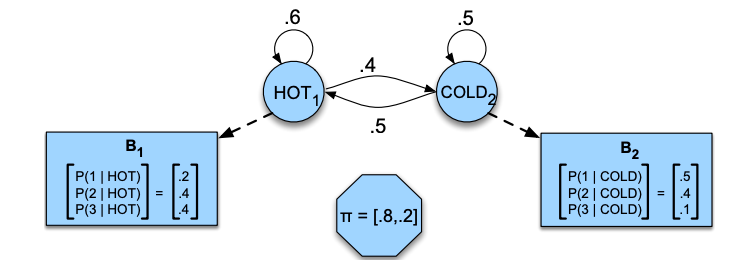}
                \caption{Hidden Markov Model example}
             \end{center}
        \end{figure}
    
    \subsubsection{Recognition using HMMs}
        Each and every test letter is treated as a new observed sate and we try to compute the hidden state, i.e., the most likely character for the observed state. \\
            To compute the hidden state the Hidden Markov Model makes use of three different probabilities:
            \begin{enumerate}
                \item \textit{Initial Probabilities} \\
                Initial probability is the probability of the test character being the first character of any statement. 
                \item \textit{Transition Probabilities} \\
                Transition probabilities are the probabilities of transitioning from one character to another, i.e., if the test character is \emph{'a'} what is probability of \emph{'a'} transition to \emph{'b', 'c'} ans so on.
                \item \textit{Emission Probabilities} \\
                Emission probability is the probability of the test character under consideration representing a particular English language character. Naive Bayes' Theorem is used to compute these probabilities. \\ \\
            \end{enumerate}
            
        After computing all the above probabilities its time for character recognition ans we carried out character recognition in two different ways:
        \begin{itemize}
            \item \textit{A Simplified Approach} \\
                The algorithm used in this approach rather than using all the three probabilities used only two of the three and those are the emission probabilities and the initial probabilities of the characters. As we are not considering the transition probabilities so, there is no connection between two different hidden states. The final character that is being assigned to the test character is solely based on the emission probability and the probability of occurrence of that very character.  
            \item \textit{Viterbi Algorithm} \\
                This algorithm finds the most likely sequence of the hidden states also known as the \textit{Viterbi Path}. It considers all three probabilities ans would store the most likely previous character and the probability of transitioning from the previous character to the current character times the emission probability of the current character. And once the entire probability table is populated using Viterbi algorithm we shall backtrack and find the most likely state.
            \end{itemize}
\begin{table*}[h!]
    \centering
    \caption{OCR accuracy using all the three modes for sample text lines extracted from a single document. The first column says number of characters in the text line / number of words}
    \begin{tabular}{ ||p{2.5cm}||p{2.5cm}||p{1cm}||p{2.5cm}||p{1cm}||p{2.5cm}||p{1cm}||}
    \hline
     \hline
     \multirow{2}{2.5cm}{Number of characters / words} & \multicolumn{2}{|c|}{Horizontal Mode} & \multicolumn{2}{|c|}{Vertical Mode} &
     \multicolumn{2}{|c|}{Pass Mode}\\
     \hline
     & Number of characters correctly recognized & Accuracy & Number of characters correctly recognized & Accuracy& Number of characters correctly recognized & Accuracy\\
     \hline
     14/4   & 0    & 0 & 5  & 35.71 & 13    &92.85\\
     \hline
     17/4 &   1  & 5.88 &   3  & 17.64 &   16  & 94.11 \\
     \hline
     34/6  & 0 & 0 &4 & 11.76  &34 & 100  \\
     \hline
     35/7   & 2 & 5.71 &7 & 20&34 & 97.14\\
     \hline
     40/9  &   1  & 2.5 &   8  & 20 &   38  & 95\\
     \hline
     40/7  & 3  & 7.5 & 12  & 30  & 37  & 92.5  \\
     \hline
     46/8  & 1  & 2.17 & 7  & 15.21 & 43  & 93.47 \\
     \hline
     50/10  & 0  & 0 & 9  & 18  & 48  & 96\\
     \hline
     50/11  & 0  & 0 & 14  & 28 & 48  & 96\\
     \hline
     51/9  & 0  & 0 & 6  & 11.76 & 48  & 94.11\\
     \hline
     55/14  & 0  & 0 & 7 & 12.72 & 48  & 87.27\\
     \hline
     Average & - & 2.26\% & - & 20.07 \% & - & 94.40\%\\
     \hline
     \hline
     \end{tabular}
     
     \label{Table : 1}
\end{table*}

\section{Experiments \& Results}
    To check the efficiency of the above discussed model we conducted various experiments. In all the experiments random sequences of words were used. For our experiment we assumed the text size is of size 25 x 16 pixels. Apart from the known text size we also assume the test character set only contains lowercase/uppercase 26 English alphabets, 10 digits, and seven special symbols/punctuation marks $().,!?"$.\\
    If our test character set have $n$ characters so, we shall have $n$ observed states $O = o_1, o_2, o_3, . . , o_n$ and $n$ hidden states $H = h_1, h_2, h_3, . . , h_n$. On the basis of this we would compute : \\
    \begin{center}
        \Large{P($h_1, h_2, h_3, . . , h_n | o_1, o_2, o_3, . . , o_n$)} \qquad \small--(i)
    \end{center}
    The above \textit{eq(i)} can be re-written using Bayes' Theorem from the data that we shall use to train the system and then using the inferences \textit{(probabilistic inferences)} we shall try to guess the most appropriate character that can be assigned to given test character. \\ 
    The experimental results of the OCR on a single document image using all the three different modes are given in Table I. From the table it is clear that pass modes are giving good accuracy than other horizontal and vertical mode.

\section{Conclusion}
    In this work we have studied a system that is capable of extracting machine printed text directly from the compressed images. The feature set that gave the best results out of the all three modes in the pass mode, which are extracted from the image compressed using CCITT \textit{(The International Telegraph and Telephone Consultative Committee)} group 3 two-dimensional algorithm. A probabilistic model based on Hidden Markov Models is used to recognize the text using the feature set.\\
    There are a lot of things which if addressed can improve the efficiency of the given model like: taking into consideration the grammar and frequently used patterns of the language, improving the capability of the system so that it can recognize text having different fonts and sizes.\\
    Finally the efficiency of the model studied in this work is clearly lower than that of the systems that works on uncompressed images but there is definitely a potential in the presented model for further enhancing the performance and potential applications of the systems may be huge in the fields where large images/documents are stored and/or transmitted.

\end{document}